# Ball Mill Fault Prediction Based on Deep Convolutional Auto-Encoding Network

**X.K. Ai[1], K. Liu[2], W. Zheng[1]\*, Y.G. Fan[2]\*, X.W. Wu[2], P.L. Zhang[1], L.Y. Wang[1], J.F. Zhu[1], Y. Pan [1]**

[1]State Key Laboratory of Advanced Electromagnetic Technology, International Joint Research Laboratory of Magnetic Confinement Fusion and Plasma Physics, School of Electrical and Electronic Engineering, Huazhong University of Science and Technology, Wuhan 430074, China
[2]Wugang Resources Group Co., Ltd. Ezhou 436000, China

Corresponding author: W. Zheng, Y.G. Fan
E-mail: zhengwei@hust.edu.cn, fanygwrgc@126.com

**Abstract**

Ball mills play a critical role in modern mining operations, making their bearing failures a significant concern due to the potential loss of production efficiency and economic consequences. This paper presents an anomaly detection method based on Deep Convolutional Auto-encoding Neural Networks (DCAN) for addressing the issue of ball mill bearing fault detection. The proposed approach leverages vibration data collected during normal operation for training, overcoming challenges such as labeling issues and data imbalance often encountered in supervised learning methods. DCAN includes the modules of convolutional feature extraction and transposed convolutional feature reconstruction, demonstrating exceptional capabilities in signal processing and feature extraction. Additionally, the paper describes the practical deployment of the DCAN-based anomaly detection model for bearing fault detection, utilizing data from the ball mill bearings of Wuhan Iron & Steel Resources Group and fault data from NASA's bearing vibration dataset. Experimental results validate the DCAN model's reliability in recognizing fault vibration patterns. This method holds promise for enhancing bearing fault detection efficiency, reducing production interruptions, and lowering maintenance costs.

## 1. Introduction

Ball mills play a pivotal role in modern mining operations, widely employed in crucial processes such as mineral crushing and grinding[1],[2]. Bearings within ball mills are critical components responsible for supporting and rolling rotors to ensure smooth machine operation. The structure of bearings usually includes inner rings, outer rings, rolling elements, and cages[3]. Typically operating under prolonged high-speed rotation, ball mill bearings endure harsh working conditions, making them prone to issues like wear, fatigue, and even fractures. These bearing failures have a profound impact on ball mill performance. Once a bearing failure occurs, it leads to reduced efficiency, lower production rates, and potential equipment downtime. Delayed detection or neglect of these failures can escalate maintenance and replacement costs, resulting in substantial economic losses for enterprises. Therefore, the mining industry has a vested interest in bearing fault detection and early warning systems.

Current methods for bearing fault detection primarily rely on sensors to collect data from ball mill bearings, such as vibration, temperature and so on. Subsequently, rule-based algorithms are established based on the fault mechanisms and operational patterns derived from this data to achieve fault detection. For example, real-time vibration waveforms collected from bearings are analyzed using techniques like Fast Fourier Transform (FFT) analysis, Haar Wavelets, cepstrum, amplitude or frequency demodulation analysis, among others [4]-[10], to assess bearing conditions. However, rule-based fault detection methods have limitations. Ball mill bearings can experience various fault types, necessitating extensive engineering effort to create warning rule algorithms for each fault type. Some faults are even related to specific ball mill structural dimensions, operating parameters, or unexpected human errors, making the fault patterns complex and challenging to deduce their underlying principles. Moreover, rule-based algorithms lack robustness when encountering new or previously unconsidered faults. Updating algorithms necessitate the expertise of human analysts to discern their underlying principles and subsequently manually incorporate relevant fault warning rules into the system. This heavy reliance on human resources further underscores the need for more automated and robust fault detection methodologies.

With the rapid development of machine learning technologies, data-driven machine learning methods have provided new solutions for ball mill fault detection. These methods leverage extensive data from different operational states of bearings to autonomously learn fault mechanisms, overcoming the limitations of rule-based algorithms. Within machine learning, supervised learning and anomaly detection represent two primary branches. Supervised learning models based on algorithms such as Support Vector Machines and Convolutional Nural Networks have been applied to bearing fault detection [11]-[14], yielding good performance. However, supervised learning models require labeled datasets of normal and faulty data for training and demand a relatively balanced distribution of these two classes of data. In practice, obtaining fault precursor samples is prohibitively costly since faults can result in equipment damage. Moreover, severe equipment damage is rare due to regular maintenance, resulting in very limited fault precursor data. This leads to extreme data imbalance in supervised learning methods, negatively impacting model performance[15],[16]. Additionally, the diverse causes of faults make it challenging to construct universal rule-based algorithms for labeling a large number of fault precursor samples, often relying on manual annotation by professionals, thereby increasing labor costs. As a result, supervised learning-based fault prediction methods have limitations.

In contrast, anomaly detection method is an emerging machine learning approach that exclusively utilizes data from normal device operation for training, addressing challenges related to data labeling and class imbalance. However, some prevalent anomaly detection algorithms, , such as Autoencoder[17], One-Class Support Vector Machine[18], K Nearest Neighbors[19], Angle-Based Outlier Detection[20], lack the ability to automatically extract features from high-dimensional raw input samples. Directly using high-dimensional raw samples for training and inference results in massive computation and extremely time-consuming processes. And manual feature extraction may lead to the loss of critical information in the data. This paper presents an anomaly detection predictor based on a Deep Convolutional Autoencoder Neural Network (DCAN) for fault detection in ball mill bearings. This network is trained and tested using vibration waveform samples collected from ball mill bearing accelerometers. The structure of the neural network includes the modules of convolutional feature extraction and transposed convolutional feature reconstruction, enabling the model to automatically learn vibration waveform features without manual definition, effectively capturing complex fault patterns.

Subsequently, We establish a ball mill fault prediction system based on DCAN predictor and deploy it on the ball mills of Wuhan Iron & Steel Resources Group. In the deployment experiments, the DCAN model successfully reconstructs vibration samples from normal bearing operation. As no faults occurred during the deployment period, and there were no historical fault data records, we generated simulated abnormal vibration samples by altering the amplitude, frequency, and harmonic composition of normal signals. Test results demonstrate that the model effectively distinguishes abnormal samples with waveforms different from normal ones. Finally, we train and conduct inference on the DCAN model using publicly available

NASA bearing vibration fault samples, further validating the reliability of the DCAN-based fault predictor in discriminating bearing fault vibration waveforms.

This paper is organized as follows: Section 2 introduces the design of Deep Convolutional Autoencoding Networks and the bearing fault predictor based on anomaly detection. Section 3 presents ball mill fault prediction system based on DCAN predictor and its deployment results on ball mill of Wuhan Iron & Steel Resources Group. Section 4 illustrates the performance of the DCAN predictor in recognizing fault samples from NASA bearings. Section 5 summarizes the content of the article.

## 2. Design of bearing fault predictor

In this section, we delve into the design of a fault predictor for ball mill bearings. The vibration sensor employed in this study utilizes a 485S accelerometer to capture vibration data. It samples the monitoring bearings at 1024 Hz every 10 minutes, accumulating data for 4 seconds before ceasing collection. Importantly, the sensor records vibrations in three directions: longitudinal, lateral, and axial (x, y, z axes) of the bearing. Consequently, each sampling event yields three one-dimensional arrays, each containing 4096 sample points. These three directional samples are then concatenated to form a 3×4096 two-dimensional dataset, serving as input samples for our predictive model. This novel dataset structure efficiently encapsulates the temporal and spatial vibration characteristics of the ball mill bearing. In light of the unique properties of our sampled data, this section introduces an anomaly detection method based on a deep convolutional auto-encoding network for real-time fault prediction in ball mill bearings. Section 2.1 elucidates the structure of the deep convolutional auto-encoding neural network, while Section 2.2 describes the fault prediction process for ball mill bearings.

### 2.1 Deep convolutional auto-encoding network

The structure of the deep convolutional auto-encoding neural network is illustrated in Figure 1. The neural network autonomously learns vibration waveform features, eliminating the need for manual feature definition and effectively capturing complex fault patterns. Its structure comprises three key modules: the convolutional feature extraction module, the feature auto-encoding module, and the transposed convolutional feature reconstruction module.

In the convolutional feature extraction module, input samples pass through three layers of two-dimensional convolutional layers, where each layer's two-dimensional kernels slide in the time and space directions for feature extraction. It outputs an output feature map with dimensions of 16×1×57. This output is then flattened into a one-dimensional array of length 912, constituting the extracted feature data.

The feature auto-encoding module compresses and subsequently reconstructs these extracted features. It takes a one-dimensional array of length 916 as input and employs five layers of fully connected layers for compression and reconstruction. The hidden layers have a length of 200, and the LeakyReLU activation function is utilized to mitigate issues such as gradient vanishing and overfitting, enabling the construction of a deeper network[20].

The transposed convolutional feature reconstruction module reshapes the reconstructed one-dimensional array into a 16×1×57 input feature map. It then undergoes upsampling through three layers of two-dimensional transposed convolutional layers to restore it to a two-dimensional array of the same size as the input sample, i.e., the reconstructed sample.

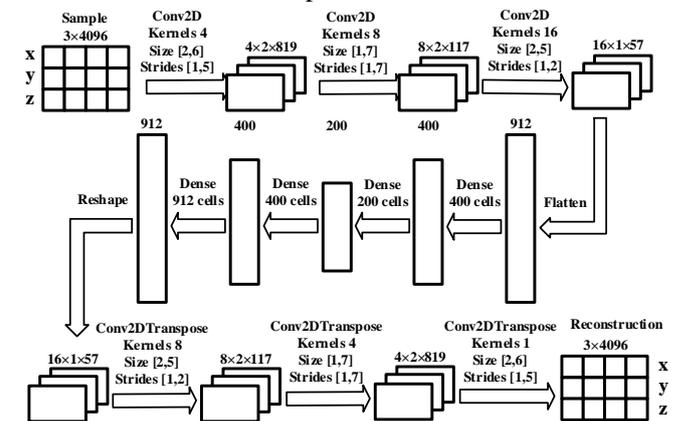

Figure 1 The structure of the deep convolutional auto-encoding neural network

### 2.2 Bearing fault predictor based on anomaly detection

When this auto-encoder model is employed for anomaly detection, the mean squared error (MSE) is used to measure the similarity between input samples and reconstructed samples. During training, MSE serves as the loss function and utilizes only normal historical samples for model training. For real-time fault detection, the mode evaluates the data samples collected by the sensor and computes the MSE. The model exhibits good reconstruction performance for normal samples, the MSE values for these are relatively small. Conversely, the reconstruction performance for untrained anomaly samples is subpar, leading to larger MSE values. When the anomaly score surpasses a predefined threshold, the system initiates a fault warning. The workflow of the fault predictor for ball mill bearing is depicted in Figure 2.

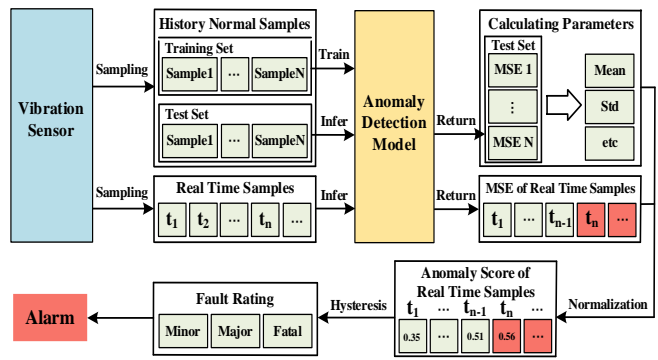

Figure 2 The workflow of the fault predictor based on anomaly detection for ball mill bearing

A ball mill is a device composed of multiple bearing components, so fault warning of the ball mill requires multiple fault predictors to monitor critical bearing locations. However, the amplitude of vibration waveforms and background noise levels may slightly differ for bearings in different positions during actual operation. This variation can lead to differences in the MSE range of normal samples returned by detectors at different monitoring sites. Furthermore, for the same fault, the MSE disparity between normal and fault samples may not be uniform across different positions. These factors can affect personnel's comprehensive assessment of the overall state of the ball mill and complicate the selection of warning thresholds. To address this issue, the fault predictors normalize the returned MSE to obtain an anomaly score. To compute normalization parameters, a set of normal samples is chosen as a test dataset, and various parameters such as the mean and variance of the MSE returned by each sample in the test dataset are calculated before inference.

When the anomaly score exceeds the preset threshold, the system triggers a fault warning. The warning threshold is categorized into three levels representing low, medium, and high-level anomaly alarms, characterizing the severity of anomalies in the samples. To mitigate the impact of interference-induced false alarms during normal operation in practical engineering, the fault determination incorporates a hysteresis control strategy. Specifically, the hysteresis alarm strategy sets a time window comprising 30 sample points, with the last sample point corresponding to the current moment. When the anomaly score of real-time samples exceeds the threshold and no fault alarms have occurred within the time window before the current moment, a fault alarm is triggered if the number of anomalous samples within the time window exceeds 16. If fault alarms have already occurred within the time window before the current moment, the anomaly detector becomes more sensitive, triggering a fault alarm when the number of anomalous samples within the time window exceeds 12.

## 3. Deployment on Ball Mill

As the ball mill consists of multiple bearing components, we design the ball mill fault prediction system that integrates multiple DCAN-based fault predictors, which can monitor the status of multiple critical parts of the ball mill in real-time. The workflow of the ball mill fault prediction system deployed on the devices of Wuhan Iron & Steel Resources Group is shown in Figure 3. Six fault predictors are deployed on a ball mill, located on both sides of the three bearing areas of the electric motor, transmission gear, and grinding cylinder. When monitoring the status of the ball mill in real-time, the raw data collected by the vibration sensor will automatically be fed into the corresponding fault predictor. At each sampling time, the anomly scores, alarm levels, and other information returned by each predictor will be considered as a status report and stored in the database. When the fault occurs, the system can locate the fault based on the location of the alarm, and visualize the fault reports on the monitoring website, making it easy for users to formulate relevant maintenance strategies. During model training, the system workflow follows the green arrows. The historical raw data is retrieved from the database to train the predictors. And the predictors calculate the mean and variance of the MSEs returned from the normal sample test set, which are used for MSE normalization in real-time analysis.

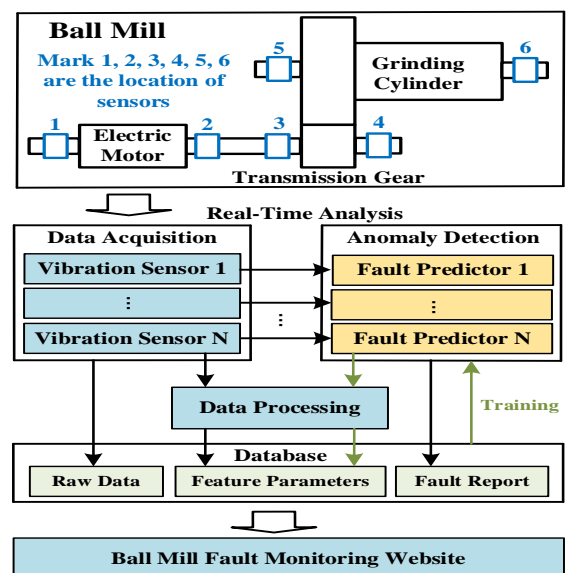

Figure 3 The workflow of the ball mill fault prediction system

Since March 2023, the DCAN-based ball mill fault prediction system has been operational on the ball mills at Wuhan Iron & Steel Resources Group, providing real-time fault warnings. In this section, we analyze the deployment performance of the bearing fault predictor installed in transmission gear of a ball mill and validate whether the DCAN model can effectively distinguish fault vibration signals from normal ones. Before April 10th, we trained the DCAN model using 10000 randomly selected normal samples. Starting from April 10th, the trained model commenced monitoring the bearings and issuing real-time fault warnings.

During real-time fault warnings, DCAN compresses and reconstructs normal samples, as depicted in Figure 4. In Figure 4(a), a lateral (x-axis) vibration signal from a randomly selected normal sample on 2023-4-15 11:18 is displayed. The spectrum of this signal, presented in Figure 4(b), illustrates that under normal operating conditions, the main frequency of the lateral vibration signal is 136 Hz, with secondary frequencies distributed around 60 Hz and some harmonic components in the high-frequency range. Figure 4(c) compares the input signal waveform with the model-reconstructed signal, showing a close resemblance with $MSE_x = 0.286$. Figure 4(d) illustrates the distribution of $MSE_x$ during normal bearing operation, primarily falling in the range of 0.2 to 0.4 for normal samples.

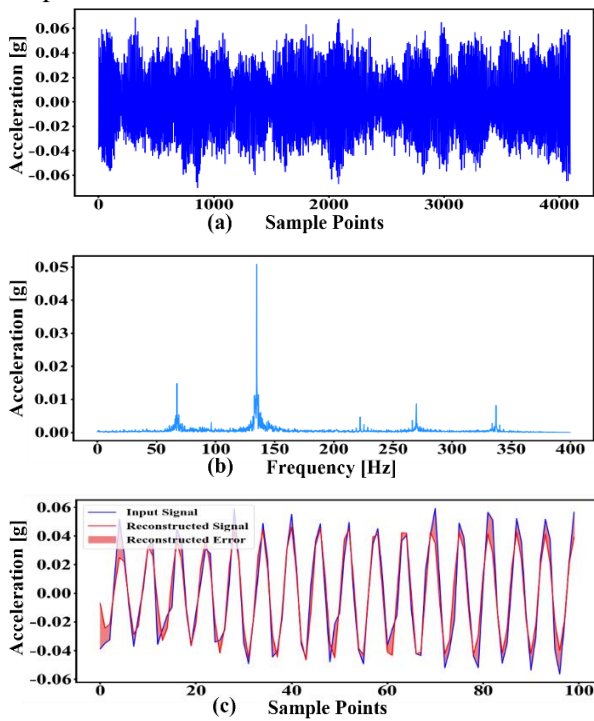

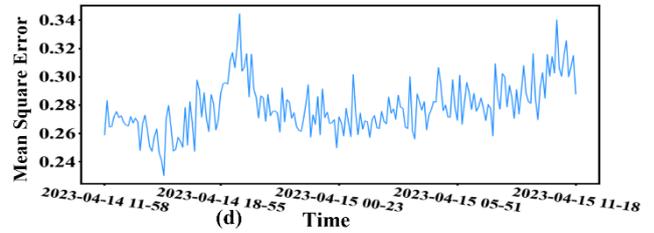

Figure 4 Reconstruction performance of DCAN on normal samples from ball mill bearing

For bearing fault samples, as no faults occurred during deployment and no historical fault samples were available, we simulated abnormal vibration waveform samples for testing the model. In Section 4, our analysis of real bearing vibration signals revealed significant differences in amplitude, frequency, and harmonic composition between fault signals and normal signals. Therefore, this section attempts to simulate abnormal vibration signals by altering the amplitude, frequency, and harmonic composition of normal signals from the ball mill bearing. Below are the abnormal signals generated by adjusting the lateral signals of the previously mentioned normal samples for model testing.

*Altering signal's frequency*

By altering the time scale of normal signals, frequency abnormal signals are obtained. We compressed the time scale of normal signals to half their original size, the abnormal signal is depicted in Figure 5(a) as the blue line. The frequencies of the abnormal signal's harmonics are twice those of the normal signal, with the main frequency $f_m = 272 \text{Hz}$. The red line represents the model-reconstructed signal, which exhibits significant differences from the blue line, with $MSE_x = 3.015$. In Figure 5(b), the abnormal signal was generated by stretching the normal signal by a factor of two, halving the frequencies of its harmonics, and setting the main frequency $f_m = 68 \text{Hz}$. The red line, representing the model-reconstructed signal, also shows noticeable differences from the blue line, with $MSE_x = 2.746$.

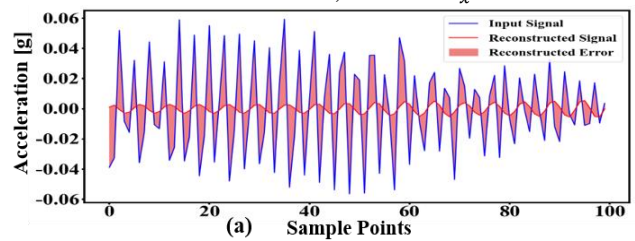

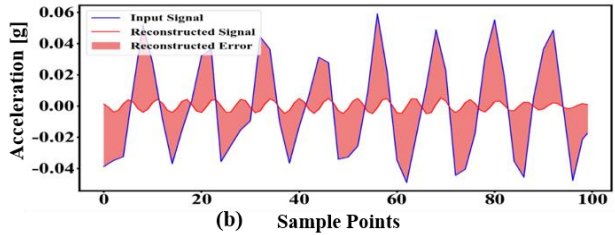

Figure 5 Reconstruction performance of DCAN on frequency abnormal samples

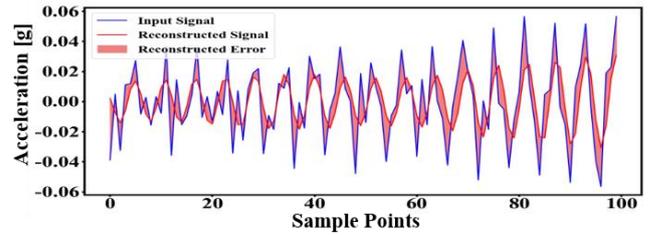

Figure 7 Reconstruction performance of DCAN for abnormal sample incorporating interference signal

*Adding interference signal*

Adding interference signal to normal signal to synthesize abnormal signal. As shown in Figure 6, a sawtooth wave with a frequency of 136Hz and a peak value of 0.04 g was superimposed as an interference signal to synthesize the abnormal signal. According to fourier transform theory, this interference signal contains the fundamental 136Hz and multiple specific frequency harmonic components. After superimposing the interference signal on the normal signal, the frequency domain introduces the harmonics of the interference signal, making the abnormal signal distinct from the normal signal in the spectrum. Furthermore, the fundamental frequency of the interference signal matches the main frequency of the normal signal, and when they are superimposed, it allows for the adjustment of the amplitudes of the main components of the normal signal.. When the synthesized abnormal signal is input to the model for reconstruction, Figure 7 displays a local comparison of the two signals, with $MSE_x = 2.991$.

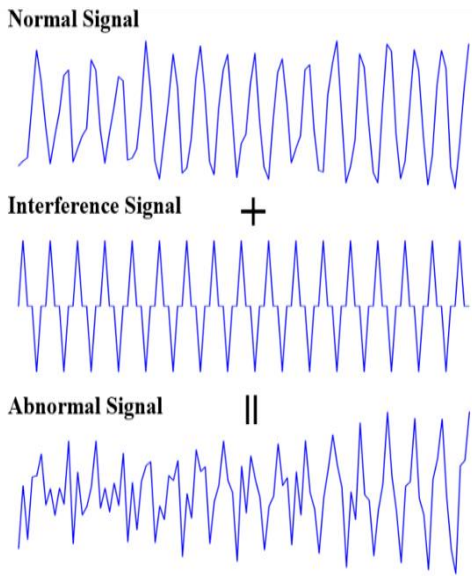

Figure 6 Adding interference signal to normal signal to synthesize abnormal signal.

By comparing the experimental results, we find that the DCAN model effectively reconstructs normal signals, with $MSE_x$ values for all evaluated normal signals primarily falling in the range of 0.2 to 0.4. However, for various simulated abnormal signals, the model's reconstruction performance is poor, resulting in $MSE_x$ values greater than 2, signifying significant differences from normal signals. The model evaluation in the longitudinal and axial directions is consistent with the lateral experimental results. In summary, the experimental results demonstrate that the DCAN model can effectively distinguish abnormal samples with waveforms differing from normal signals. This underscores the potential of our approach for real-time fault detection in ball mill bearings.

## 4. Validation with NASA Bearing Database

In order to comprehensively validate the reliability and performance of the DCAN-based bearing fault predictor in discerning bearing fault vibration waveforms, extensive tests were conducted using publicly available NASA bearing vibration fault samples[22]. Detailed information regarding the NASA bearing database can be found on the website [NASA Bearing Dataset | Kaggle](). This database provides valuable vibration data on rotating bearings over three distinct time periods, namely Set 1, Set 2, and Set 3. Multiple sensors monitor the rotating bearings in each time period, with each sensor referred to as a channel (Ch). The database encompasses data for three distinct types of bearing faults: inner race defect, outer race failure, and roller element defect. In Set 1, an inner race defect emerged towards the end of the data collected in Ch 5, while a roller element defect emerged at the end of data collection in Ch 7. In Set 2, an outer race failure emerged towards the end of the data collected in Ch 1. For the experiments presented in this section, data from these three channels were selected as the test set. To construct the training set, we combined the remaining normal samples from the three Sets and randomly selected 30,000 samples.

Upon completion of model training, inference was performed on the data from the three selected channels within the test set. Figure 8 shows the Reconstruction performance of DCAN on normal samples from NASA bearing. It is known that the Ch1 channel of Set 2 had normal samples from 2004-2-12 to 2004-2-17. The MSE values returned by the model for these samples are shown in Figure 8(a), with MSE primarily distributed below 0.8. Figure 8(b) illustrates the waveform of a randomly selected normal sample from that period, comprising 4096 sampling points. Figure 8(c) presents the spectrum of this normal sample, indicating that during normal operation, the vibration signal's main frequency is 158 Hz, with some low-frequency harmonic components. Figure 8(d) shows a comparison between this normal signal and the model-reconstructed signal, highlighting a high degree of similarity, with $MSE = 0.656$.

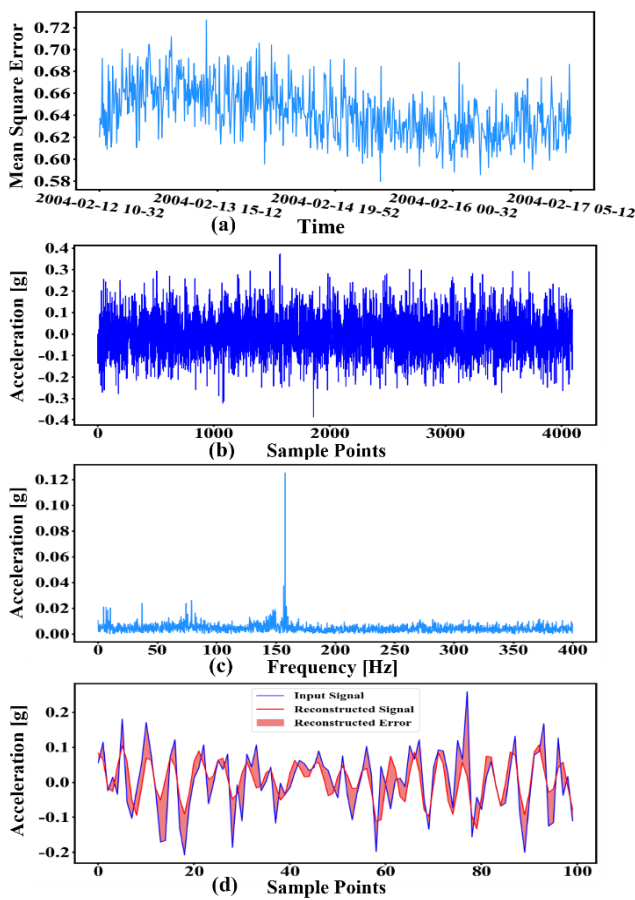

Figure 8 Reconstruction performance of DCAN on normal samples from NASA bearing database

*Inner Race Defect*

Figure 9 shows the reconstruction performance of the DCAN model on samples of inner race defect. Figure 9(a) shows the MSE values of all samples from Ch 5 of Set 1 returned by the model. The MSE curve initially maintains low values, primarily below 0.8, but later exhibits a significant increase. As the inner race defect occurred towards the end of the data collection, this suggests that the MSE values for samples during the fault period are notably higher than those for preceding normal samples, underscoring the model's ability to discern inner race defect. A sample from the fault period, indicated by the red line, was randomly selected for analysis. Its spectrum, as shown in Figure 9(b), displays a substantial DC component, a reduced amplitude at $f = 158\text{Hz}$, and prominent harmonics in the $f < 50\text{Hz}$ range and near $f = 190\text{Hz}$. The harmonic composition of the fault sample significantly differs from that of normal samples. Figure 9(c) presents a local comparison between the fault sample and its reconstructed counterpart, revealing relatively poor waveform reconstruction, with $MSE = 4.212$.

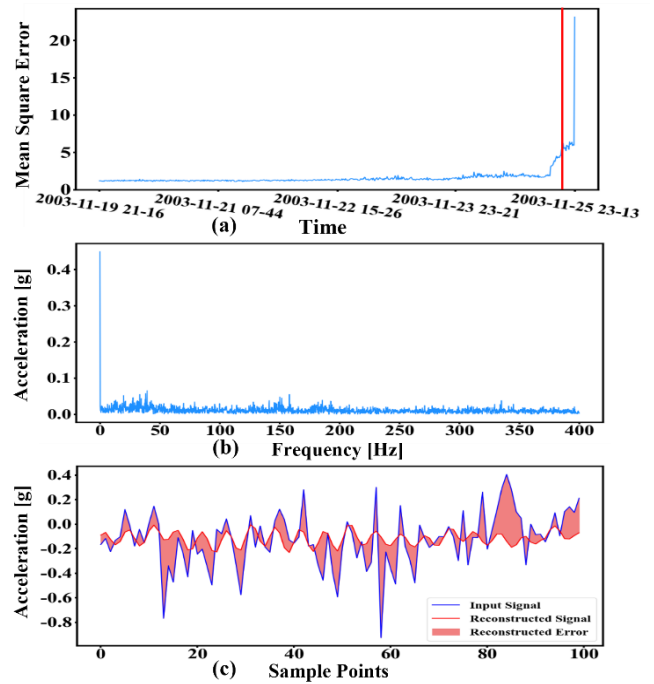

Figure 9 Reconstruction performance of the DCAN model on samples of inner race defect from NASA bearing database

*Outer Race Failure*

Figure 10 shows the reconstruction performance of the DCAN model on samples of outer race failure. Figure 10(a) shows the MSE values of all samples from Ch 1 of Set 2 returned by the model. The MSE curve exhibits a significant increase at the end. As the outer race failure occurred towards the end of the data collection, this suggests the model's capability to distinguish the outer race failure. A sample from the fault period, indicated by the red line, was randomly selected for analysis. Its spectrum, as shown in Figure 10(b), reveals a main frequency near 100 Hz, with a more complex harmonic composition compared to normal signal spectra. Figure 10(c) presents a local comparison between the fault sample and its reconstructed counterpart, demonstrating relatively poor waveform reconstruction, with $MSE = 2.239$.

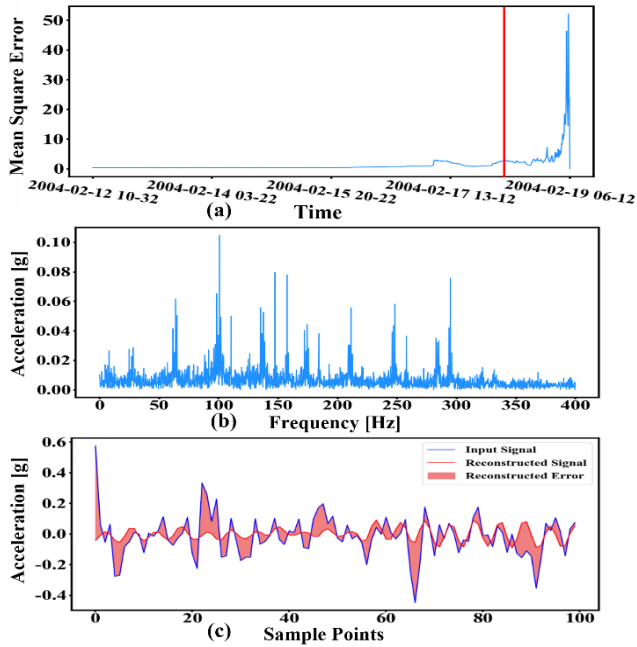

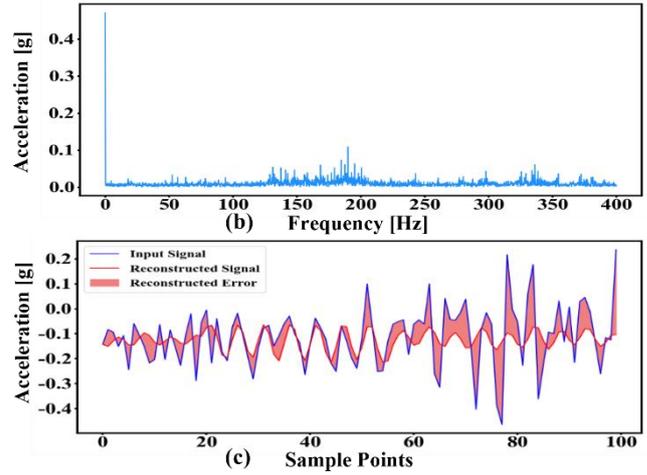

Figure 11 Reconstruction performance of the DCAN model on samples of roller element defect from NASA bearing database

## 5. Summary

This study aimed to address the issue of bearing fault prediction in modern mining mills, focusing on ball mill bearings. We designed an anomaly detection predictor based on the Deep Convolutional Auto-Encoding Network (DCAN) for bearing fault detection in ball mills. The DCAN-based model is trained and applied for inference using vibration waveform samples collected by acceleration sensors on ball mill bearings. The structure of the neural network includes the modules of convolutional feature extraction and transposed convolutional feature reconstruction, enabling it to autonomously learn the features of raw signals and capture complex fault patterns more effectively. The DCAN-based predictor discriminates between normal and fault samples by evaluating the similarity between input and reconstructed samples. This article provides comprehensive insights into the deployment and practical application of the DCAN-based predictor for bearing fault detection. We outline the development of a ball mill fault prediction system based on DCAN predictors, which was successfully deployed in the ball mills of Wuhan Iron & Steel Resources Group. Experimental data is drawn from ball mill bearings within the Wuhan Iron & Steel Resources Group and bearing vibration fault samples from NASA for experimentation, validating the DCAN model's performance across different datasets. Through model training and inference, we have successfully demonstrated the DCAN model's capability to distinguish between normal and fault samples, thereby confirming its reliability in recognizing fault vibration waveforms. Particularly for the ball mill bearings at Wuhan Iron & Steel Resources Group, we have presented the DCAN model's reconstruction performance on various vibration signal samples. By comparing normal and abnormal samples, we have shown that the DCAN model excels at reconstructing

Figure 10 Reconstruction performance of the DCAN model on samples of outer race failure from NASA bearing database

*Roller Element Defect*

Figure 11 shows the reconstruction performance of the DCAN model on samples of roller element defect. Figure 11(a) shows the MSE values of all samples from Ch 7 of Set 1 returned by the model. The MSE curve exhibits a significant increase at the end. As the roller element defects occurred towards the end of the data collection, this suggests that the model can distinguish the roller element defect. A sample from the fault period, indicated by the red line, was randomly selected for analysis. Its spectrum, as shown in Figure 11(b), displays a substantial DC component, a main frequency near 190 Hz, and a more complex harmonic composition in the frequency domain compared to normal signals. Figure 11 (c) presents a local comparison between the fault sample and its reconstructed counterpart, revealing relatively poor waveform reconstruction, with $MSE = 1.678$.

In summary, the DCAN model effectively distinguishes between normal and fault samples in NASA bearings. The model exhibits good reconstruction performance for normal samples but poorer performance for fault samples. This experiment confirms the reliability of the DCAN model in recognizing fault vibration waveforms in bearings.

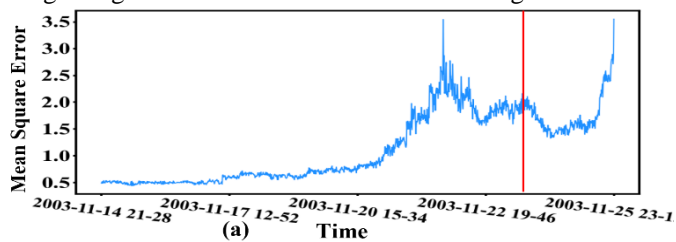

normal signals but exhibits poorer reconstruction performance on abnormal signals, providing strong support for bearing fault detection. Additionally, further validation was conducted using NASA's bearing vibration fault samples, encompassing various fault types such as inner and outer race defects and rolling element defects. Through training and testing on these datasets, the results illustrate the DCAN model's ability to identify different fault types, further validating its reliability.

In conclusion, this research significantly advances the field of bearing fault detection, showcasing the applicability of Deep Convolutional Auto-Encoding Networks. This study equips industry professionals with valuable methods and tools to enhance the reliability and safety of industrial equipment. With the continuous evolution of deep learning technology, we anticipate further breakthroughs and broader applications for the DCAN anomaly detection method in the future.

## Acknowledgement

The authors are very grateful for the help of J-TEXT team and Wugang Resources Group Co., Ltd. This work is supported by the Wugang Resources Group Investment Technology Renovation Project under Grant SBZNWHGJXT and by the National Natural Science Foundation of China under Grant No.51821005.

## References

[1] Bor, A. Jargalsaikhan, B. Lee, J. et al. Effect of Different Milling Media for Surface Coating on the Copper Powder Using Two Kinds of Ball Mills with Discrete Element Method Simulation. 2020.

[2] Zhao, X Z. Shaw L. Modeling and Analysis of High-Energy Ball Milling Through Attritors. Metallurgical and Materials Transactions A. 48, 4324–4333. 2017.

[3] Zhou, J. Qin, Y. et al. Fault detection of rolling bearing based on FFT and classification. Journal of Advanced Mechanical Design Systems and Manufacturing. Vol.9, No.5. 2015.

[4] Bediaga, I. Mendizabal, X. et al. Ball Bearing Damage Detection Using Traditional Signal Processing Algorithms. IEEE Instrumentation & Measurement Magazine. Vol.16, No.2. 2013.

[5] Lipovszky, G. Sólyomvári, K. Varga, G. Vibration Testing of Machines and their Maintenance. publications of the american jewish historical society, 1990.

[6] Lin, H C. Ye, Y C. Reviews of bearing vibration measurement using fast Fourier transform and enhanced fast Fourier transform algorithms. Advances in Mechanical Engineering. 11(1). 2019.

[7] Kahaei, M H. Torbatian, M. Poshtan, J. Detection of Bearing Faults Using Haar Wavelets. Ieice Transactions on Fundamentals of Electronics Communications & Computer Sciences. 89(3): 757-763, 2006.

[8] Harris, T A. Crecelius, W J. et al. Rolling Bearing Analysis. Journal of Tribology. 1986.

[9] Devaney, M J. Eren, L. Detecting motor bearing faults. IEEE Instrum. Meas. Mag. Vol. 7, No. 4, pp. 30-50. 2004.

[10] Ho, D. Randall, R B. Optimization of bearing diagnostics techniques using simulated and actual bearing fault signals. Mechanical Systems & Signal Processing. 14(5):763-788. 2000.

[11] Pandarakone, S E. Mizuno, Y. Nakamura, H. A Comparative Study between Machine Learning Algorithm and Artificial Intelligence Neural Network in Detecting Minor Bearing Fault of Induction Motors.Energies. 12. 2019.

[12] Ye, R. Wang, W J. et al. Bearing Fault Detection Based on Convolutional Self-Attention Mechanism. 2020 IEEE 2nd International Conference on Civil Aviation Safety and Information Technology. 2020.

[13] Li, B. Goddu, G. Chow, M Y. Detection of common motor bearing faults using frequency-domain vibration signals and a neural network based approach[C]//American Control Conference.IEEE, 2002.

[14] Pan, T. Chen, J. Zhou, Z. et al. A Novel Deep Learning Network via Multiscale Inner Product With Locally Connected Feature Extraction for Intelligent Fault Detection. IEEE Transactions on Industrial Informatics. 2019:5119-5128. 2019.

[15] Jang, J. et al. Feature Concentration for Supervised and Semi-supervised Learning with Unbalanced Datasets in Visual Inspection. IEEE Transactions on Industrial Electronics. PP(99):1-1. 2020.

[16] Nguyen, M. H. Impacts of Unbalanced Test Data on the Evaluation of Classification Methods. International Journal of Advanced Computer Science & Applications. 10(3):497-502. 2019.

[17] Hinton, G. E. Autoencoders, minimum description length and Helmholtz free energy. Advances in Neural Information Processing Systems San Mateo, 1994, 6.

[18] Schölkopf. et al. Estimating the Support of a High-Dimensional Distribution. Neural Computation, 2001.

[19] Su, M. Y. Real-time anomaly detection systems for Denial-of-Service attacks by weighted k-nearest-neighbor classifiers. Expert Systems with Applications, 2011, 38(4):3492-3498.

[20] Kriegel, H. P. et al. Angle-based outlier detection in high-dimensional data// Proceedings of the 14th ACM SIGKDD International Conference on Knowledge Discovery and Data Mining, Las Vegas, Nevada, USA, August 24-27, 2008. ACM.

[21] El Mellouki O, Khedher MI, El-Yacoubi MA, et al. Abstract Layer for LeakyReLU for Neural Network Verification Based on Abstract Interpretation. IEEE ACCESS, 2023(33401-33413).

[22] Qiu, H. Lee, J. Lin, J. et al. Wavelet filter-based weak signature detection method and its application on rolling element bearing prognostics. Journal of Sound & Vibration. 289(4-5):1066-1090. 2006.